\documentclass[11pt,a4paper]{article}
\usepackage[hyperref]{acl2020}
\usepackage{times}
\usepackage{latexsym}
\usepackage{textcomp}

\usepackage{microtype}
\usepackage{amsmath}
\usepackage{arydshln}
\usepackage{multirow}
\usepackage{graphics}
\usepackage{comment}
\usepackage{url}
\usepackage{textcomp}
\usepackage{mathtools}

\aclfinalcopy 
\def\aclpaperid{1891}

\title{Contrastive Self-Supervised Learning for Commonsense Reasoning}

\author{{Tassilo Klein \hspace{1cm} Moin Nabi} \\
SAP AI Research, Berlin, Germany\\
{\tt $\{$tassilo.klein, m.nabi$\}$@sap.com}}
\date{}

\begin{document}
\maketitle
\begin{abstract}

We propose a self-supervised method to solve \emph{Pronoun Disambiguation} and \emph{Winograd Schema Challenge} problems.
Our approach exploits the characteristic structure of training corpora related to so-called ``trigger'' words, which are responsible for flipping the answer in pronoun disambiguation.  
We achieve such commonsense reasoning by constructing pair-wise contrastive auxiliary predictions. To this end, we leverage a \emph{mutual exclusive loss} regularized by a \emph{contrastive} margin.
Our architecture is based on the recently introduced transformer networks, BERT, that exhibits strong performance on many NLP benchmarks. Empirical results show that our method alleviates the limitation of current supervised approaches for commonsense reasoning. This study opens up avenues for exploiting inexpensive self-supervision to achieve performance gain in commonsense reasoning tasks. \footnote{Code available at \url{https://github.com/SAP-samples/acl2020-commonsense/}}

\end{abstract}

\section{Introduction}

Natural language representation learning (e.g., BERT~\cite{devlin2018bert}, etc.) can capture rich semantics from text and consistently improve the performance of downstream natural language processing (NLP) tasks. However, despite the recent progress, the task of \emph{commonsense reasoning} is still far from being solved. Among many factors, this can be attributed to the strong correlation between attainable accuracy and training corpora size and quality. A particular case in point is the Winograd Schema Challenge (WSC)~\cite{levesque2012winograd}. Despite its seeming simplicity for humans, it is still not solved by current algorithms.

Below is a popular example of a question-answer pair from the binary-choice pronoun coreference problem \cite{lee2017end} of WSC: \\
\\
\emph{\textbf{Sentence-1:}} \emph{The trophy doesn\textquotesingle t fit in the suitcase because {\textbf{it}} is too \underline{small}.}\\ 
\emph{\textbf{Answers:}} \emph{\textbf{A)}} the trophy \emph{\textbf{B)}} the suitcase\\ \\
\emph{\textbf{Sentence-2:}} \emph{The trophy doesn\textquotesingle t fit in the suitcase because {\textbf{it}} is too \underline{big}.}\\ 
\emph{\textbf{Answers:}} \emph{\textbf{A)}} the trophy \emph{\textbf{B)}} the suitcase 
\\
\\
For humans resolving the pronoun ``it'' to ``the suitcase'' is straightforward. However, a system without
the capacity of commonsense reasoning is unable to conceptualize the inherent relationship and, therefore, unable to distinguish ``the suitcase'' from the alternative ``the trophy''.

Recently, the research community has experienced an abundance in methods proposing to utilize latest word embedding and language model (LM) technologies 
for commonsense reasoning~\cite{kocijan19acl, he2019hybrid, ye2019align, ruan2019exploring, trinh2018simple, klein-nabi-2019-attention}. The underlying assumption of these methods is that, since such models are learned on large text corpora (such as Wikipedia), they implicitly capture to a certain degree commonsense knowledge. As a result, models permit reasoning about complex relationships between entities at inference time.
Most of the methods proposed a two-stage learning pipeline. They are starting from an initial self-supervised model, commonsense-aware word embeddings are then obtained in a subsequent fine-tuning (ft) phase. Fine-tuning enforces the learned embedding to solve the downstream WSC task only as a plain co-reference resolution task.

However, solving this task requires more than just employing a language model learned from large text corpora.
We hypothesize that the current self-supervised pre-training tasks (such as \emph{next sentence prediction}, \emph{masked language model}, etc.) used in the word embedding phase are too ``easy'' to enforce the model to capture commonsense. Consequently, the supervised fine-tuning stage is not sufficient nor adequate for learning to reason commonsense. This is particularly more severe when pre-training on commonsense-underrepresented corpora such as Wikipedia, where the authors often skip incorporating such information in the text, due to the assumed triviality. In this case, the supervised fine-tuning does not seem to be enough to solve the task, and can only learn to ``artificially'' resolve the pronoun based on superficial cues such as dataset and language biases~\cite{trichelair2018evaluation,DBLP:journals/corr/abs-1810-00521,trichelair-etal-2019-reasonable,emami2019knowref, kavumba2019choosing}.

In this work, we propose to use minimal existing supervision for learning a commonsense-aware representation.
Specifically, we provide the model with a supervision level identical to the test time of the Winograd challenge. For that, we introduce a self-supervised pre-training task, which only requires pair of sentences that differ in as few as one word (namely, ``trigger'' words). It should be noted that the notion of trigger words is inherent to the concept of Winograd Schema questions. Trigger words are responsible for switching the correct answer choice between the questions. In the above example, the adjectives big and small act as such trigger words. 
Given the context established by the trigger word, candidate answer A is either right in the first sentence and wrong in the second, or vice-versa. As is evident from the example, trigger words give rise to the mutual-exclusive relationship of the training pairs. The proposed approach targets to incorporate this pairwise relationship as the only supervisory signal during the training phase. 
Training in such a contrastive self-supervised manner is inducing a commonsense-aware inductive bias. This can be attributed to several factors. Optimization enforces the classifier to be more rigorous in its decision as well as consistent across pairs while being discriminative. Specifically, in the absence of strong individual sentence signals, the model seeks to combine weak signals across pairs. This unsupervised task is much harder to learn compared to the supervised task, and resolving the respective associations requires a notion of commonsense knowledge. Consequently, we postulate that training with contrastive self-supervised fashion allows for learning more in-depth word relationships that provide better generalization properties for commonsense reasoning. 

For that, we propose to incorporate a Mutual Exclusive (MEx) loss~\cite{NIPS2016_6333} during the representation learning phase by maximizing the mutual exclusive probability of the two plausible candidates. Specifically, given a pair of training sentence, the pronoun to be resolved is masked out from the sentence, and the language model is used to predict such only one of the candidates can fill in the place of masked pronoun while fulfilling the mutual-exclusivity condition.
In this self-supervised task, the labels (i.e., correct candidates) do not have to be known a priori. Thus it allows learning in an unsupervised manner by exploiting the fact that the data is provided in a pairwise fashion. 

Our contributions are two-fold: \textbf{(i)} we propose a novel self-supervised learning task for training commonsense-aware representation in a minimally supervised fashion. \textbf{(ii)} we introduce a pair level mutual-exclusive loss to enforce commonsense knowledge during representation learning.

\begin{figure*}[t!]
\begin{center}
\includegraphics[width=.92\textwidth]{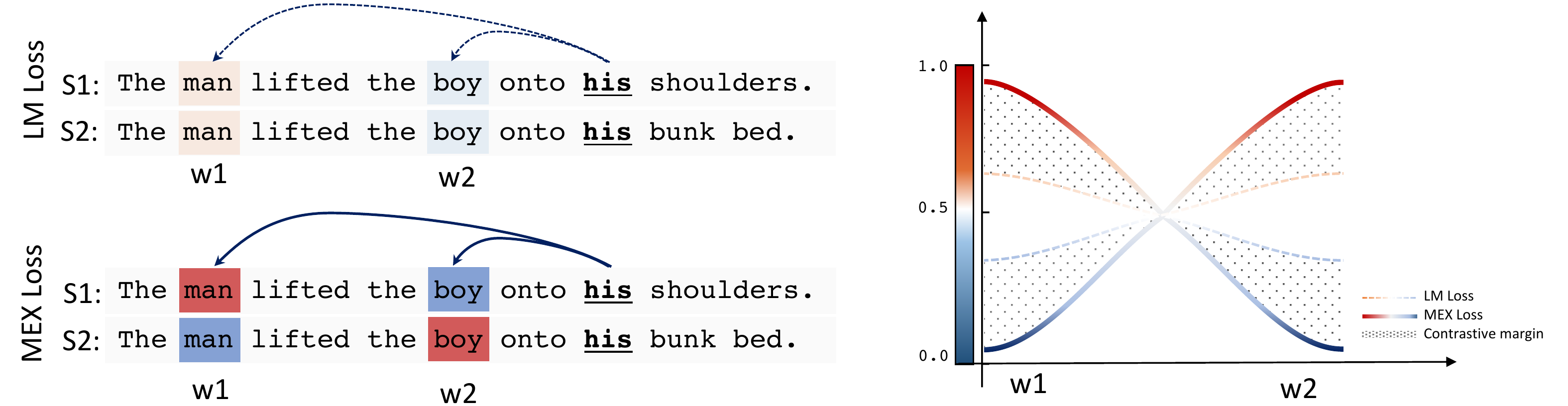}
\end{center}
\caption{Contrastive Self-supervised Learning for a particular sentence. Colors show the likelihood of different words. Weak commonsense signal manifests in the likelihood of both candidates to be around 0.5 for the LM-only loss (shown in dash lines); incorporating the MEx loss (shown in solid lines) leverages mutual exclusivity of the candidates, enforcing the classifier to be more rigorous and consistent across pairs (best shown in color).}\label{fig:fig_1}
\end{figure*}

\section{Previous Works}
There is a wealth of literature on commonsense reasoning, but we only discuss here the ones most related to our work and refer the reader to the recent analysis paper by \cite{trichelair-etal-2019-reasonable}.

Traditional attempts on commonsense reasoning usually involve heavy utilization of annotated knowledge bases (KB), rule-based reasoning, or hand-crafted features \cite{bailey2015winograd, schuller2014tackling, sharma2015towards}. Only very recently and after the success of natural language representation learning, several works proposed to use \emph{supervised learning} to discover commonsense relationships, achieving state-of-the-art in multiple benchmarks (see, e.g., \cite{kocijan19acl, he2019hybrid, ye2019align, ruan2019exploring}). As an example, \cite{kocijan19acl} has proposed to exploit the labels for commonsense reasoning directly and showed that the performance of multiple language models on Winograd consistently and robustly improves when fine-tuned on a similar pronoun disambiguation problem dataset. Despite the success of these methods, we posit that \emph{unsupervised learning} is still more attractive for commonsense reasoning tasks, because curating a labeled dataset entailing all existing commonsense is likely to be an unattainable objective. Very recently, unsupervised learning has also been applied successfully to improve commonsense reasoning in a few works ~\cite{trinh2018simple, klein-nabi-2019-attention}. The most pioneering work in this space is probably by ~\cite{trinh2018simple}, where the authors proposed to use BERT as a (pseudo) language model to compute the likelihood of candidates replacing the pronoun, and the corresponding ratio giving rise to answer. In another recent work, \cite{klein-nabi-2019-attention} proposed a metric based on the maximum attention score for commonsense reasoning. While these papers show that BERT can implicitly learn to establish complex relationships between entities, our results suggest that solving commonsense reasoning tasks require more than unsupervised models learned from massive text corpora. 
We note that our model is different from all of the methods above. A key difference is that they require fine-tuning, or explicit substitution or heuristic-based rules, whereas our method learns a commonsense-aware representation in self-supervised fashion.

\section{Contrastive Self-supervised Reasoning}
The goal of the proposed approach is to exploit the mutual-exclusive nature of the training samples of commonsense reasoning corpora.  
Given two sentences where the only difference between them is the trigger word(s), we postulate that the pairwise pronoun disambiguation is mutually exclusive. We formulate this idea using a contrastive loss and use this to update the language model. The proposed contrastive loss decomposes into two components:
\begin{equation}
\mathcal{L}(f_\theta) = \mathcal{L}(f_\theta)_{MEx} + \mathcal{L}(f_\theta)_{CM}
\end{equation}
Here $f$ is the language model parameterized by $\theta$.
The first term, $\mathcal{L}_{MEx}$ enforces the \underline{M}utual \underline{Ex}clusivity of the answers across pairs. As such, it is a relaxation of the Exclusive-OR (XOR) operator w.r.t. candidates. The second term, $\mathcal{L}_{CM}$ constitutes the \underline{C}ontrastive \underline{M}argin. It enforces a margin between the candidate likelihoods from the language model. Whereas $\mathcal{L}_{MEx}$ operates across pairs, $\mathcal{L}_{CM}$ considers the candidates of each pair. Although both terms encourage the same property (mutual exclusivity of the answers), we empirically observed that adding ${CM}$ increases stability. It should be noted that the proposed approach does not make use of any class label information explicitly. Rather, it solely exploits the structural information of the data. 
In terms of the language model, we leverage BERT for Masked Token Prediction~\cite{devlin2018bert}. This entails replacing the pronoun by a mask, i.e., \emph{[MASK]}. As a result, we yield probabilities for the candidates of each sentence.
\\
\noindent\textbf{Preliminaries: } Given an associated pair of training sentences, i.e., $\left(s_j, s_{j+1}\right)$, where the difference between the sentence pairs are the trigger words. Let $c_i$ and $c_{i+1}$ be the two answer candidates for the masked pronoun resolution task. Then employing BERT for Masked Token Prediction~\cite{devlin2018bert} provides $p\left(c_i|s_j\right)$ and $p\left(c_{i+1}|s_j\right)$, i.e., the likelihood of the first and the second candidate being true in sentence $s_j$, respectively. It should be noted, if a candidate consists of several tokens, the corresponding number of \emph{[MASK]} tokens is used in the masked sentence. The candidate probability then corresponds to the average of log-probabilities of each composing token.

Since a candidate cannot be the right answer for the first and second sentence in the pair, we yield a logical term that holds \emph{true} for viable answers. It is worth noting that the logical expression is not unique as many logical equivalents exist:
\begin{equation}
\label{eq:binarylogical}
\left(c_{i,1} \oplus c_{i+1,1}\right)\land\left(c_{i,2} \oplus c_{i+1,2}\right) \land \left(c_{i,1} \oplus c_{i,2}\right)
\end{equation}
Here $\oplus$ denotes the XOR operator and $\mathbf{c}_{i,j}\in\{0,1\}$ denotes the binary state variable corresponding to candidate $\mathbf{c}_{i}$ in sentence $\mathbf{s}_j$.

\noindent\textbf{Mutual-Exclusive Loss:}
In order to be differentiable, the discrete logical term of Eq.~\ref{eq:binarylogical} has to be converted into a ``soft'' version. To this end, we replace the binary variables with their corresponding probabilities.  Similarly, the logical operators are replaced accordingly to accommodate for the probabilistic equivalent. \\
With $a\oplus b=\left(a\land\neg b\right) \lor \left(\neg a \land b\right)$ a logical decomposition of the XOR operator, we adopt the following replacement scheme:  \textbf{(i)}  $\bigwedge_i^k x_i$ is replaced by $\prod_i^k x_i$, \textbf{(ii)} $\bigvee_i^k x_i$ is replaced by $\sum_i^k x_i$, \textbf{(iii)} the not operation of a binary variable $\neg x_i$ is replaced by $1-x_i$. Thus, transforming all the logical terms of Eq.~\ref{eq:binarylogical}, we yield the following soft-loss equivalent:
\begin{multline}
    \mathcal{L}_{MEx}=-\gamma\sum_{i=i+2,}^{N}\mathbf{p}_{i,1} \mathbf{p}_{i+1,2}\left(1-\mathbf{p}_{i,2}  \mathbf{p}_{i+1,1}\right)\\
    +\mathbf{p}_{i,2} \mathbf{p}_{i+1,1}\left(1-\mathbf{p}_{i,1}\mathbf{p}_{i+1,2}\right)
\end{multline}
Here $\mathbf{p}_{i,j}=p\left(c_{i}|s_{j}\right)\in\left[0,1\right]$ denotes the probability of candidate $\mathbf{c}_{i}$ being the right answer in sentence $\mathbf{s}_j$, $\gamma$ is a hyperparameter, and $N$ corresponds to the number of training samples.
Intuitively speaking, as no labels are provided to the model during training, the model seeks to make the answer probabilities less ambiguous, i.e., approximate binary constitution. As the model is forced to leverage the pairwise relationship in order to resolve the ambiguity, it needs to generalize w.r.t. commonsense relationships. As such, the task is inherently more challenging compared to, e.g., supervised cross-entropy minimization.

\noindent\textbf{Contrastive Margin: }
In order to stabilize optimization and speed-up convergence, it is beneficial to augment the MEx loss with some form of regularization. To this end, we add a contrastive margin. It seeks to maximize difference between the \emph{individual} candidate probabilities of the language model and is defined as,
\begin{equation}
\begin{split}
    \mathcal{L}_{CM}=-\alpha \cdot \max\left(0,|p_{i,j}-p_{i,j+1}|+\beta\right),
\end{split}
\label{eq:margin}
\end{equation}
with $\alpha, \beta$ being hyperparameters. See Fig.~\ref{fig:fig_1} for a schematic illustration of the proposed method.

\section{Experiment \& Results}
In this work, we use the PyTorch~\cite{Wolf2019HuggingFacesTS} implementation of BERT. Specifically, we employ a pre-trained BERT \emph{large-uncased} architecture. The model is trained for $25$ epochs using a batch size of $4$ (pairs), hyperparameters $\alpha=0.05$, $\beta=0.02$ and $\gamma=60.0$, and Adam optimizer at a learning rate of $10^{-5}$. We approach commonsense reasoning by first fine-tuning the pre-trained BERT LM model on the DPR training set~\cite{rahman-ng-2012-resolving}. Subsequently, we evaluate the performance on four different tasks.

\noindent\textbf{Pronoun Disambiguation Problem:}
The first evaluation task is on PDP-60~\cite{davis2016human}, which aims the pronoun disambiguation. As can be seen in Tab.~\ref{tab:results} (top), our method outperforms all previous unsupervised results by a significant margin of at least (+15.0\%). Next, we have the alternative approaches making use of a supervisory signal during training. Here, our method outperforms even the best system  (78.3\%) by (+11.7\%).

\noindent\textbf{Winograd Schema Challenge:}
The second task is WSC-273~\cite{levesque2012winograd}, which is known to be more challenging than PDP-60. Here, our method outperforms the current \emph{unsupervised} state-of-the-art~\cite{trinh2018simple} (62.6\%), as shown in Tab.~\ref{tab:results} (middle). Specifically, our method achieves an accuracy of (69.6\%), which is (+7\%) above the previous best result. Simultaneously, the proposed approach is just slightly lower than the best \emph{supervised} approach~~\cite{kocijan19acl}.

\noindent\textbf{Definite Pronoun
Resolution:}
The third task is DPR~\cite{rahman-ng-2012-resolving}, which resembles WSC. Compared to the latter, it is significantly larger in size. However, according to ~\cite{trichelair2018evaluation}, it is less challenging due to several inherent biases. Here the proposed approach outperforms the best alternative by a margin of (+3.7\%), as can be seen in Tab.~\ref{tab:results} (lower part).

\noindent\textbf{KnowRef:}
The fourth task is KnowRef~\cite{emami2019knowref}, which is a coreference corpus tailored to remove gender and number cues. The proposed approach outperforms the best alternative by a margin of (+4.5\%), as can be seen in Tab.~\ref{tab:results} (bottom).

\noindent\textbf{Ablation study on contrastive margin:}
The contrastive margin term was incorporated in our method as a regularizer, mainly for the sake of having faster convergence. As such, discarding it during optimization has a minor impact on the accuracy of most benchmarks (less than 1\% on WSC, DPR, KnowRef). However, on PDP, we noticed a wider margin of more than 10\%. 

\begin{table}[ht!]
\begin{center}
\centering
\begin{tabular}{ll}
\hline
\multicolumn{2}{c}{\textbf{PDP-60} \texttt{(sup.)}~\cite{davis2016human}} \\
\hline
Patric Dhondt (WS Challenge 2016) & 45.0 \% \\
Nicos Issak (WS Challenge 2016) & 48.3 \% \\
Quan Liu (WS Challenge 2016-\emph{winner}) & 58.3 \% \\
USSM + Supervised DeepNet & 53.3 \% \\
USSM + Supervised DeepNet + 3 KB & 66.7 \% \\
BERT-ft~\cite{kocijan19acl} & 78.3 \% \\
\hdashline
\multicolumn{2}{c}{PDP-60 \texttt{(unsupervised)}} \\
\hdashline
Unsupervised Sem. Similarity (USSM) & 55.0 \% \\ 
Transformer LM \cite{vaswani2017attention} & 58.3 \% \\
BERT LM \cite{trinh2018simple}& 60.0 \% \\
MAS \cite{klein-nabi-2019-attention} & 68.3 \%\\
DSSM \cite{wang2019unsupervised} & 75.0 \%\\
\hline
\textbf{CSS (Proposed Method)} & \textbf{90.0 \%}\\
\hline\hline
\multicolumn{2}{c}{\textbf{WSC-273} \texttt{(sup.)}~\cite{levesque2012winograd}} \\
\hline
USSM + KB & 52.0\% \\
USSM + Supervised DeepNet + KB & 52.8 \% \\
Transformer ~\cite{vaswani2017attention} & 54.1 \% \\
Know. Hunter \cite{emami2018knowledge}& 57.1 \% \\
GPT-ft~\cite{kocijan19acl} & 67.4 \% \\
BERT-ft~\cite{kocijan19acl} & 71.4 \% \\
\hdashline
\multicolumn{2}{c}{WSC-273 \texttt{(unsupervised)}} \\
\hdashline
Single LMs~\cite{trinh2018simple} & 54.5 \% \\
MAS~\cite{klein-nabi-2019-attention} & 60.3 \% \\
DSSM \cite{wang2019unsupervised} & 63.0 \%\\
Ensemble LMs~\cite{trinh2018simple} & 63.8 \% \\
\hline
\textbf{CSS (Proposed Method)} & \textbf{69.6 \%}\\
\hline\hline
\multicolumn{2}{c}{\textbf{DPR}~\cite{rahman-ng-2012-resolving}}\\
\hline
\cite{rahman-ng-2012-resolving} & 73.0\% \\
\cite{peng-etal-2015-solving} & 76.4 \% \\
\hline
\textbf{CSS (Proposed Method)} & \textbf{80.1 \%}\\
\hline
\hline
\multicolumn{2}{c}{\textbf{KnowRef}~\cite{emami2019knowref}} \\
\hline
E2E~\cite{emami2019knowref}  & 58.0 \% \\
BERT-ft~\cite{emami2019knowref} & 61.0 \% \\
\hline
\textbf{CSS (Proposed Method)} & \textbf{65.5 \%}\\
\hline
\end{tabular}
\caption{Results on different tasks. From Top to bottom: PDP, WSC, DPR, KnowRef. The first two task performances are subdivided into two parts. Upper part: supervised, lower part: unsupervised. 
}
\label{tab:results}
\end{center}
\end{table}

\section{Discussion}
In contrast to supervised learning, where semantics is directly injected through ``labels'', the self-supervised-learning paradigm avoids labels by employing a pre-text task and exploits the structural ``prior'' of data as a supervisory signal. In this paper, this prior corresponds to the Winograd-structured twin-question pairs, and the pre-text task is to switch the correct answer choice between the pairs using ``trigger'' words. We postulate that training in such a contrastive self-supervised manner allows for learning more commonsense-aware word relationships that provide better generalization properties for commonsense reasoning. We acknowledge that this prior is strong in terms of data curation, i.e., expert-crafted twin pairs. However, during training, we provide the model to have access to a supervision level equal to the test time, i.e., not making use of the labels. 
Therefore, maximizing the mutual exclusive probability of the two plausible candidates 
is inducing a commonsense-aware inductive bias without using any label information and by merely exploiting the contrastive structure of the task itself. This is confirmed by our approach, reaching the performance of the most recent supervised approaches on multiple benchmarks. 
At last, we note that our model is different from the self-supervised contrastive learning methodology in \cite{chen2020simple}, which focuses on learning powerful representations in the self-supervised setting through batch contrastive loss. A key difference compared to this method is that they generate the contrastive pairs as data augmentations of given samples, whereas in our setting the auxiliary task of ``mutual exclusivity'' is enforced on given contrastive pairs.

\section{Conclusion}
The proposed approach outperforms all approaches on PDP and DPR tasks. At the more challenging WSC task, it outperforms all unsupervised approaches while being comparable in performance to the most recent supervised approaches. Additionally, it is less susceptible to gender and number biases as the performance on KnowRef suggests. All this taken together confirms that self-supervision is possible for commonsense reasoning tasks. We believe in order to solve commonsense reasoning truly, algorithms should refrain from using labeled data, instead exploit the structure of the task itself. Therefore, future work will aim at relaxing the prior of Winograd-structured twin-question pairs. Possibilities are automatically generating an extensive collection of similar sentences or pre-training in a self-supervised fashion on large-scale Winograd-structured datasets, such as the recently published WinoGrande~\cite{sakaguchi2019winogrande}.  Furthermore, we seek to investigate the transferability of the obtained inductive bias to other commonsense-demanding downstream tasks, which are distinct from the Winograd-structure.

\bibliography{acl2020}
\bibliographystyle{acl_natbib}
\end{document}